%% file: main.tex
\documentclass[10pt,twocolumn,letterpaper]{article}

\usepackage[pagenumbers]{cvpr} 

\input{preamble}
\definecolor{cvprblue}{rgb}{0.21,0.49,0.74}
\usepackage[pagebackref,breaklinks,colorlinks,allcolors=cvprblue]{hyperref}

\usepackage{booktabs}
\usepackage{multirow}
\usepackage{graphicx}
\usepackage[table]{xcolor}

\definecolor{proprietarybg}{HTML}{DCEFEA}
\definecolor{opensourcebg}{HTML}{E2EEFA}
\definecolor{reasoningbg}{HTML}{E9E4F5}
\definecolor{ourmodelsbg}{HTML}{F9DDC8}
\definecolor{ourrowbg}{HTML}{D5E9FA}

\def\confName{CVPR}
\def\confYear{2027}

\title{Tool-to-Self: Internalizing Visual Tool Capabilities via Scaffolded Reinforcement Learning}

\author{
    \textbf{Jin Cui}\textsuperscript{\rm 1},
    \textbf{Hao Dong}\textsuperscript{\rm 3},
    \textbf{Xinyue Long}\textsuperscript{\rm 1,\rm 2},
    \textbf{Boran Zhao}\textsuperscript{\rm 1,\rm 2} \thanks{Corresponding author},
    \textbf{Pengju Ren}\textsuperscript{\rm 1} \vspace{4mm}\\
    \textsuperscript{\rm 1}State Key Laboratory of Human-Machine Hybrid Augmented Intelligence,\\
    and Institute of Artificial Intelligence and Robotics, Xi'an Jiaotong University\\
    \textsuperscript{\rm 2}School of Software Engineering, Xi'an Jiaotong University \\
    \textsuperscript{\rm 3}ELLIS Institute Finland and Tampere University  
}

\begin{document}
\maketitle
\input{sec/0_abstract}    
\input{sec/1_intro}
{
    \small
    \bibliographystyle{ieeenat_fullname}
    \bibliography{main}
}


\end{document}

%% file: sec/0_abstract.tex
\begin{abstract}
Vision-language models (VLMs) increasingly rely on external visual tools to access fine-grained visual evidence, but repeated tool execution incurs substantial inference overhead and deployment dependencies. Moreover, conventional tool-use RL can improve visual-task performance, yet the resulting tool gains are often unstable. Motivated by this, we propose \textit{Tool-to-Self (T2S)}, a tool-scaffolded reinforcement learning framework that turns external visual tools into temporary training scaffolds. T2S cold-starts tool-use ability and observation prediction from tool-repairable examples, then progressively anneals tool execution on a per-tool basis during RL. By combining answer-correctness and observation-consistency rewards with privileged on-policy self-distillation from actual tool observations, T2S effectively teaches the model both to reproduce tool outputs and to reason from them. Experiments across diverse visual perception and reasoning benchmarks demonstrate consistent improvements over both explicit tool-use models and implicit zooming models while enabling tool-free inference with reduced latency.

\end{abstract}

%% file: sec/1_intro.tex
\section{Introduction}
\label{sec:intro}

Vision-language models (VLMs) increasingly solve perception-heavy tasks by acting on an image rather than observing it once. Following the \textit{``thinking with images''} paradigm~\cite{openai2025thinking}, recent models learn through supervised fine-tuning and reinforcement learning (RL) to crop or zoom into relevant regions~\cite{zheng2025deepeyes,su2025pixelreasoner,lai2025minio3}, or to invoke specialist tools such as OCR engines, detectors, and chart parsers~\cite{hong2025deepeyesv2,zong2024mova}, and then reason over the returned observations. These gains, however, come with substantial inference overhead: \emph{every call incurs tool execution and visual re-encoding, lengthens the reasoning trajectory, and introduces a deployment dependency whose errors may propagate to the final answer.}  More fundamentally, the benefits of tool use do not necessarily come from the observations. MED~\cite{ma2026med} attributes over 70\% of the gains from crop-and-zoom RL to improved intrinsic, tool-free capabilities, while the learned tool policy primarily avoids harmful calls. Causal audits further reveal that returned observations often have little effect on final predictions~\cite{wang2026illusion}, and replacing returned crops with textual placeholders can achieve performance comparable to full thinking-with-images training~\cite{shao2026textcall}. These findings suggest that \emph{tools can serve as scaffolds for developing visual capabilities, yet conventional tool-use RL primarily optimizes tool invocation and final-answer rewards, without explicitly teaching models to reproduce or internalize the capabilities behind tool observations}. Consequently, such internalization remains incidental and unstable, motivating a fundamental question: \emph{\textbf{Can visual tools be used as temporary training scaffolds to transfer their capabilities into the model, rather than remaining inference-time dependencies?}}

Several recent works have begun to explore internalizing visual capabilities. Region-to-image distillation~\cite{wei2026zwz}, crop-conditioned on-policy self-distillation~\cite{visionopd2026,imagineopd2026}, and latent visual reasoning~\cite{yang2025mirage,pearl2026} transfer localized visual inspection into tool-free inference. Other approaches supervise dense perceptual targets, such as depth and segmentation~\cite{bigverdi2025aurora,qin2025covt}. However, these methods predominantly focus on crop-and-zoom or predefined perceptual targets. It remains unclear whether heterogeneous specialist capabilities, such as text recognition, object localization, and chart parsing, can also be internalized. Moreover, privileged supervision is often predefined through annotated regions, synthesized crops, or offline programs~\cite{hu2024vpd}, rather than acquired through the model's own on-policy tool requests. This leaves open how to explicitly transfer diverse tool capabilities while aligning supervision with the model's actual perceptual needs.

A practical constraint is that \textit{\textbf{internalizability is determined not by the tool itself, but by whether the information required to reproduce its observation survives the model's visual encoding.}} When decisive evidence is preserved in the encoded input, the tool primarily provides a computation for localizing, reading, or transforming information already available to the model. In contrast, when lacking that evidence, a tool operating on the original high-resolution image may expose information that the model cannot recover from its encoded input alone. Thus, the same operation can be \emph{computation-bound} or \emph{information-bound} depending on the task and input pipeline. Our analysis establishes this distinction and uses it to guide data curation: we retain observations recoverable from the encoder view and increase the input resolution when feasible, excluding examples whose decisive evidence remains inaccessible. This provides a practical scope for studying tool internalization without assuming that information lost during visual encoding can be reconstructed through training alone.

Building on this insight, we propose \textbf{Tool-to-Self (T2S)}, a scaffolded RL framework that explicitly transfers heterogeneous visual tool capabilities into the model. T2S first cold-starts tool use and observation prediction from tool-repairable examples, where actual expert observations correct the model's failed predictions. A strong external teacher then organizes these repaired trajectories into coherent tool-use supervision. During RL, T2S progressively reduces the probability of external tool execution through an environment-level annealing curriculum. When a tool call is withheld, the model must instead generate a self-simulated observation, while the actual tool is executed in a hidden auxiliary branch to provide dense reference supervision. We combine an observation consistency reward, which directly supervises the fidelity of generated observations, with privileged on-policy self-distillation (OPSD)~\cite{zhao2026opsd}, which transfers observation-conditioned reasoning from the same model's privileged branch to its tool-free policy. In this way, T2S explicitly trains the model both to reproduce tool observations and to reason effectively from them, progressively turning external tools from inference-time dependencies into training-time scaffolds.

We evaluate T2S on 13 benchmarks spanning diverse multimodal understanding and reasoning tasks. Without external tool execution, T2S outperforms most explicit tool-use and implicit zooming methods, while achieving 3.6$\times$ faster inference than tool-augmented execution. Furthermore, its advantage persists at large pass@$k$, providing evidence of genuine capability acquisition beyond mere elicitation. Our contributions are threefold:
\begin{itemize}
    \item We identify the explicit internalization of tool-provided capabilities as an overlooked learning objective in agentic visual tool use. Through controlled experiments, we show that conventional tool-use RL induces only incidental and unstable internalization, and characterize visual-encoding recoverability as a constraint on this process.

    \item We propose Tool-to-Self (T2S), a scaffolded RL framework that internalizes heterogeneous visual tool capabilities through explicit observation prediction, progressive tool withdrawal, observation consistency rewards, and privileged on-policy self-distillation.

    \item Across 13 benchmarks, T2S achieves strong tool-free performance, outperforming most explicit tool-use and implicit zooming methods with substantially lower inference cost. Pass@$k$ experiments further demonstrate capability gains beyond improved elicitation.
\end{itemize}

\begin{figure*}
    \centering
    \includegraphics[width=1\linewidth]{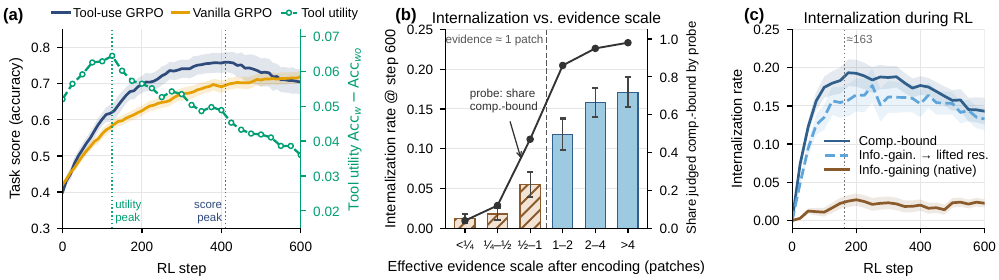}
    
    \caption{\textbf{Tool-use RL internalizes only computation-bound tool gains, and only partially.} \textbf{(a)}~Task score of tool-use and vanilla GRPO from the same SFT checkpoint on the same data (mean $\pm$ std over four runs), and tool utility of tool-use GRPO (dashed, right axis). \textbf{(b)}~Internalization rate at the final checkpoint \vs the effective scale of the decisive evidence after preprocessing; the line shows the share of questions labeled computation-bound by the encoder-view probe. \textbf{(c)}~Internalization rate during RL for computation-bound, information-gaining, and resolution-lifted information-gaining questions.}
    \label{fig:motivation}
    \vspace{-2mm}
\end{figure*}

\section{Related Work}
\label{sec:related_work}

\paragraph{Thinking with Images and Visual Tool-Use RL.}
Thinking with Images treats vision as a dynamic reasoning workspace, evolving from external tool interaction to intrinsic visual reasoning~\cite{su2025thinking}. Early approaches employ drawing and specialist perception tools~\cite{hu2024sketchpad}, programmatic image manipulation~\cite{fu2025refocus,zhang2026thyme}, or RL-trained visual interactions, including crop-and-search~\cite{zheng2025deepeyes}, pixel-space exploration~\cite{su2025pixelreasoner}, multi-turn visual search~\cite{lai2025minio3}, and heterogeneous specialist tools~\cite{hong2025deepeyesv2}. Despite their effectiveness, recent analyses reveal a gap between tool invocation and effective use of tool observations: crop-and-zoom RL gains largely arise from improved intrinsic capabilities~\cite{ma2026med}, while returned observations often have limited causal influence on final predictions~\cite{wang2026illusion}. Given the inference overhead of repeated tool execution, these findings motivate explicit internalization of tool-provided capabilities: rather than relying on increasingly complex reward shaping to stabilize tool-use RL, models can be directly supervised to reproduce and utilize tool observations, thereby reducing external tool dependencies and inference latency.

\paragraph{Intrinsic Visual Imagination and Privileged Distillation.}
Recent work internalizes visual capabilities through region-to-image distillation~\cite{wei2026zwz}, crop-conditioned on-policy self-distillation~\cite{visionopd2026,imagineopd2026}, and latent visual reasoning~\cite{yang2025mirage,pearl2026}, enabling localized inspection without external visual operations. Privileged distillation further transfers behaviors conditioned on privileged information to policies operating without it~\cite{zhao2026opsd}. T2S adopts this on-policy distillation mechanism but differs in the source and role of privileged information: rather than relying on predefined privileged signals, it acquires actual tool observations through the model's own on-policy requests. Moreover, training is complemented by explicit observation prediction and consistency supervision, while progressive tool annealing gradually transfers both observation and its utilization into the tool-free policy.

\section{Motivation}
\label{sec:motivation}

\paragraph{Tools act as a transient scaffold.}
We train the same cold-started model on the same data with tool-use GRPO, following the training settings in DeepEyesV2~\cite{hong2025deepeyesv2}, and with vanilla GRPO, which provides no tools. All other configurations are identical, and each variant is run four times. Following MED~\cite{ma2026med}, we measure tool utility as the gap between tool-available and tool-free accuracy, $\mathrm{Acc}_{w}-\mathrm{Acc}_{wo}$. As shown in Fig.~\ref{fig:motivation}a, tool-use GRPO learns faster and peaks higher ($0.76$ \vs $0.72$), but then declines, whereas vanilla GRPO keeps improving. Tool utility peaks much earlier (step 125, $0.064$) and decays to $0.036$. Since the task score keeps rising while utility shrinks, we assume the tool-free accuracy must be catching up: part of the tool's contribution may be absorbed by the model, yet nothing in the objective asks it to be retained.


\paragraph{Only computation-bound gains are internalized.}
To identify which tool gains are recoverable from the model's visual input, we introduce an \emph{encoder-view probe}. For each tool-repairable question $x\in\mathcal{D}_{\text{repair}}$, we re-run the tool on the encoder-view image $U(\Phi_B(I))$, obtained by applying the model's preprocessing under pixel budget $B$ and upsampling the result:
\begin{equation}
  x\in\mathcal{D}_{\text{comp}}\iff \hat{y}^{+}\big(T_j(U(\Phi_B(I)),c)\big)=y^{*},
  \label{eq:probe}
\end{equation}
and $x\in\mathcal{D}_{\text{info}}$ otherwise. Thus, $\mathcal{D}_{\text{comp}}$ contains questions whose tool-provided evidence remains available after encoding, whereas $\mathcal{D}_{\text{info}}$ requires information discarded by the encoder. To isolate tool-specific internalization from general RL gains, we define the \emph{internalization rate} at step $t$ as the fraction of repairable questions that the vanilla-GRPO checkpoint fails without tools, but the tool-use-GRPO checkpoint solves without tools. As shown in Fig.~\ref{fig:motivation}c, this rate on $\mathcal{D}_{\text{comp}}$ peaks at about 0.19 near step 163, shortly after the tool-utility peak, before declining to 0.14. In contrast, it remains at the noise floor of about 0.02 on $\mathcal{D}_{\text{info}}$. This sharp distinction emerges around an evidence scale of one encoder patch (Fig.~\ref{fig:motivation}b). Moreover, increasing the global resolution of information-gaining examples makes them nearly as internalizable as computation-bound ones, confirming that the barrier lies in the input pipeline rather than the questions themselves.

\paragraph{Implications.}
Incidental internalization is real but limited and unstable. T2S therefore makes it \emph{explicit}, by supervising self-simulated observations, and \emph{targeted}, by training only on observations recoverable from the model's input.


\paragraph{Computation-bound curation.}
We first apply the probe in Eq.~\eqref{eq:probe} to all repairable examples and retain those classified as $\mathcal{D}_{\text{comp}}$. For each example in $\mathcal{D}_{\text{info}}$, we progressively increase its pixel budget to the minimum feasible $B\le B_{\max}$ at which the probe classifies it as computation-bound, subject to its native resolution. Examples that remain information-gaining are discarded. For every retained example, we generate both observation targets for $\mathcal{L}_{\text{obs}}$ and reference observations for $R_{\text{obs}}$ using the corresponding encoder view $U(\Phi_B(I))$, ensuring that all supervised observations are recoverable from visual input. Finally, we balance curated dataset across tools and evidence-scale categories.

\begin{figure*}
    \centering
    \includegraphics[width=1\linewidth]{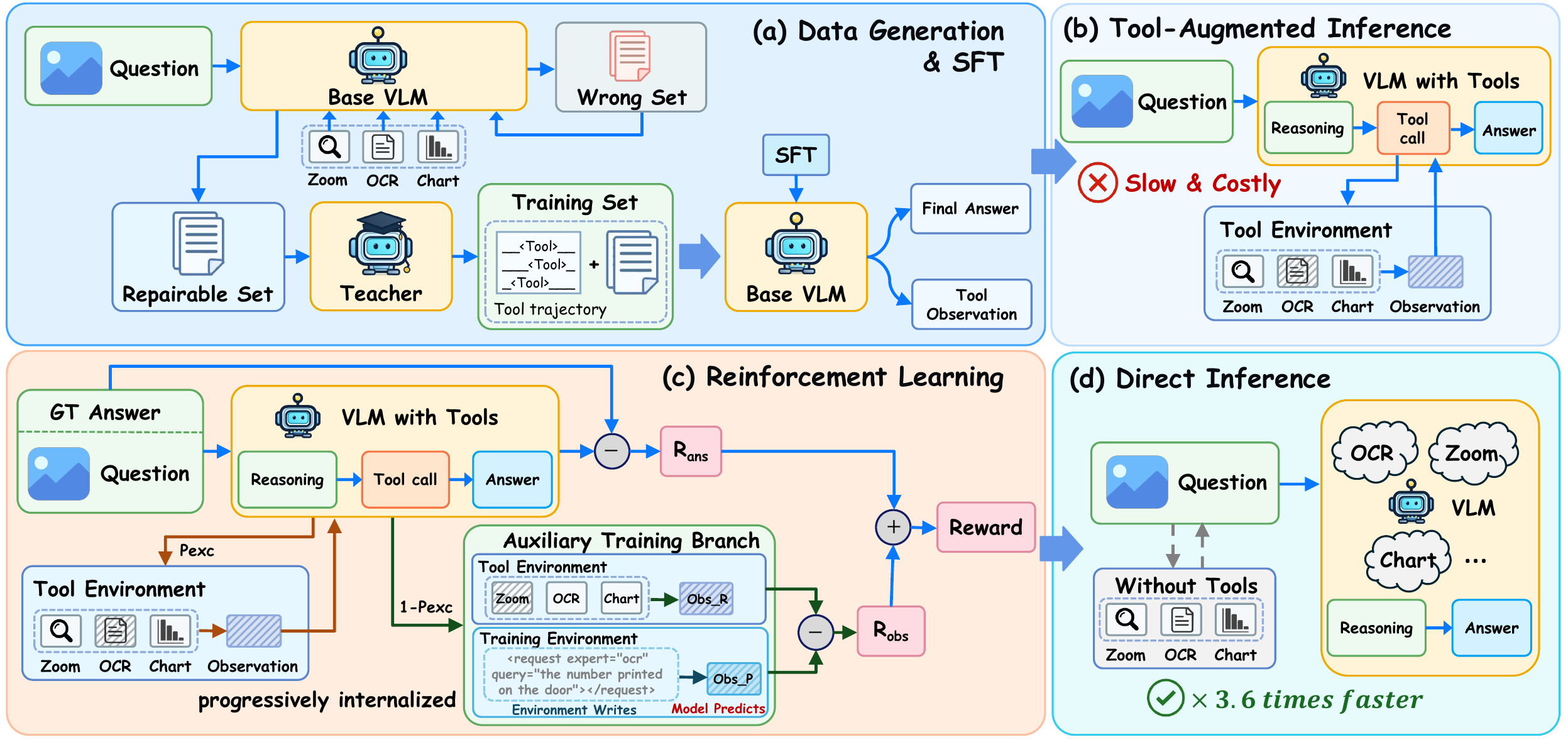}
    \caption{\textbf{Overview of T2S.} (a) Data construction and cold-start: Failed student rollouts are repaired by tool observations, and a teacher organizes successful repairs into trajectories for supervised fine-tuning. (b) Tool-augmented inference: Tool-use models depend on external execution and observation feedback. (c) Scaffolded reinforcement learning: T2S progressively replaces executed tool observations with self-simulated ones, using hidden reference observations for consistency rewards and privileged on-policy self-distillation. (d) Tool-free inference: The resulting model performs perception and reasoning without external tool execution.}
    \label{fig:main-fig}
    \vspace{-2mm}
\end{figure*}

\section{Method}
\label{sec:method}

In this section, we first review representative paradigms for supervising agentic tool use behaviors, then introduce our method and its training procedure. An overview of our proposed framework is provided in Fig.~\ref{fig:main-fig}.

\subsection{Preliminaries}
\label{sec:preliminary}

\paragraph{Reward-Engineered RL for Tool Use.}
Recent RL methods optimize on-policy trajectories using verifiable rewards for answer correctness, often augmented with tool-specific shaping terms. A schematic total reward is
\begin{equation}
\begin{aligned}
&R_{\mathrm{RL}}(\tau,x)
={}R_{\mathrm{ans}}
+\lambda_p R_{\mathrm{proto}}
+\lambda_s w(x)R_{\mathrm{succ}}\\
&+\lambda_e R_{\mathrm{eff}}
+\lambda_a R_{\mathrm{aux}}
-\mu\max(0,n_{\mathrm{tool}}-n_{\max}),
\end{aligned}
\end{equation}
where $R_{\mathrm{ans}}$ measures answer correctness, $R_{\mathrm{proto}}$ captures format and protocol validity, $w(x)R_{\mathrm{succ}}$ rewards successful tool use according to tool necessity, and $R_{\mathrm{eff}}$ measures tool effectiveness or evidence gain. $R_{\mathrm{aux}}$ summarizes other task-specific signals, while the final term penalizes excessive tool calls. These components vary across methods and are typically optimized using policy-gradient algorithms such as GRPO or GSPO.

\paragraph{On-Policy Self-Distillation (OPSD).}
OPSD mitigates distribution mismatch by matching token distributions along student-generated prefixes. The same model serves as both teacher and student: the privileged branch conditions on actual tool observations $O$, while the student relies only on the original input and its generated prefix. The objective:
{\fontsize{9.0pt}{11.4pt}\selectfont
\begin{equation}
\begin{aligned}
&\mathcal{L}_{\mathrm{OPSD}}(\theta)
= \\[-2pt]
&\mathbb{E}_{\substack{x\sim\mathcal{D}\\
\hat{\tau}\sim\pi_\theta(\cdot\mid x)}}
\Bigg[
\frac{1}{|\hat{\tau}|}
\sum_{t=1}^{|\hat{\tau}|}
D\Big(
\operatorname{sg}\big[
\pi_\theta^{\mathrm{pri}}
(\cdot\mid x,h_t,O_{\leq k(t)})
\big]
\,\Big\|\,
\pi_\theta(\cdot\mid x,h_t)
\Big)
\Bigg],
\end{aligned}
\end{equation}}

where $h_t=\hat{\tau}_{<t}$ is the student-generated prefix, $O_{\leq k(t)}$ denotes the privileged tool observations available at step $t$, $\operatorname{sg}$ stops gradients through the privileged branch, and $D$ denotes a distribution divergence. By providing dense token-level guidance on on-policy prefixes, OPSD transfers tool-conditioned behavior without relying on an external teacher.


\subsection{Problem Formulation}

Given an image--question pair $x=(I,q)$ and its ground-truth answer $y^*$, let $\mathcal{T}=\{T_j\}_{j=1}^{J}$ denote a set of visual tools. At step $t$, the model may issue a request $a_t=(j_t,c_t)$, where $j_t$ specifies the tool and $c_t$ its arguments. The tool returns an observation $o_t=T_{j_t}(I,c_t)$, which supports subsequent reasoning. Conventional tool-augmented inference relies on these external observations, whereas our goal is to internalize the recoverable computation of visual tools, enabling tool-free inference without external execution. 


\subsection{Repairable Data Construction and Cold-Start}

We construct training data in three steps. First, we run the base model without tools and collect incorrect trajectories:
\[
\mathcal{D}_{\mathrm{wrong}}
=
\{(x,\tau_s^-): \hat{y}_s^- \neq y^*\}
\]
where $\tau_s^-$ denotes the student's failed trajectory. Second, we provide the student with the corresponding expert observations $O^*$, obtained by executing the relevant visual tools, and rerun inference. We retain examples for which these observations fix the answer correctly:
\[
\mathcal{D}_{\mathrm{repair}}
=\{(x,\tau_s^-,O^*,\tau_s^+,y^*):
\hat{y}_s^- \neq y^*,\quad
\hat{y}_s^+(O^*)=y^*\}
\]
where $\tau_s^+$ denotes the corrected student trajectory conditioned on $O^*$. This criterion isolates examples in which tool observations provide actionable evidence for correcting the student's prediction. Third, we use Qwen3-VL-235B-A22B as an external teacher to organize each repairable example into a coherent tool-use trajectory $\tau^T$, comprising reasoning, tool requests, the corresponding expert observations, and the final answer. The actual tool observations are preserved as reference targets rather than regenerated by the teacher. The resulting trajectories initialize the student through supervised fine-tuning.

To explicitly supervise observation generation, we augment the standard trajectory objective with an observation-prediction loss:
\vspace{-2mm}
\begin{equation}
\mathcal{L}_{\mathrm{cold}}
=
\mathcal{L}_{\mathrm{traj}}^{\neg\mathrm{obs}}
+\alpha\mathcal{L}_{\mathrm{obs}},
\end{equation}
where $\mathcal{L}_{\mathrm{traj}}^{\neg\mathrm{obs}}$ follows the distillation objective in Sec.~\ref{sec:preliminary} but excludes observation tokens from its loss mask, and
{\fontsize{9.3pt}{11.4pt}\selectfont
\begin{equation}
\mathcal{L}_{\mathrm{obs}}
=
-\mathbb{E}_{\tau^T}
\left[
\frac{1}{|\mathcal{M}_{\mathrm{obs}}|}
\sum_{t\in\mathcal{M}_{\mathrm{obs}}}
\log\pi_\theta(\tau_t^T\mid x,\tau_{<t}^T)
\right].
\end{equation}}
Here, $\mathcal{M}_{\mathrm{obs}}$ indexes observation tokens, and $\alpha$ controls the strength of observation supervision. This objective teaches the student both to use tool observations and to predict them from its existing context.

\subsection{Scaffolded Reinforcement Learning}

\paragraph{Environment-Level Tool Annealing.}
Starting from the cold-started model, we progressively reduce external tool execution during on-policy RL. For a request to tool $j_t$, the environment samples
\vspace{-2mm}
{\fontsize{9.0pt}{11.4pt}\selectfont
\begin{equation}
e_t\sim\operatorname{Bernoulli}
\big(p_{\mathrm{exec}}^{j_t}\big),
\quad
o_t^{\mathrm{policy}}=
\begin{cases}
o_t, & e_t=1,\\
\hat{o}_t, & e_t=0,
\end{cases}
\end{equation}}
where $o_t=T_{j_t}(I,c_t)$ is the actual tool output and $\hat{o}_t$ is the model-generated observation. Each tool's execution probability is initialized at one and progressively reduced through a tool-specific curriculum. We argue it can be adjusted according to validation-set observation accuracy, reducing execution when the model becomes reliable at reproducing the corresponding observations, which is left for our future work and here we adopt a predefined annealing schedule.

When $e_t=0$, the model must continue reasoning from its self-simulated observation. The actual tool is nevertheless executed in a hidden auxiliary branch to provide a reference for training. Thus, tool withdrawal is controlled by the environment rather than an explicit penalty on tool calls.

\input{tab/main-exp}

\paragraph{Observation Consistency Reward.}
We directly evaluate the fidelity of self-simulated observations against the actual tool outputs. Let $\mathcal{M}_{\mathrm{sim}}=\{t:e_t=0\}$ denote the set of withheld calls. We define
\begin{equation}
R_{\mathrm{obs}}
=
\begin{cases}
\displaystyle
\frac{1}{|\mathcal{M}_{\mathrm{sim}}|}
\sum_{t\in\mathcal{M}_{\mathrm{sim}}}
\operatorname{sim}_{j_t}(\hat{o}_t,o_t),
& |\mathcal{M}_{\mathrm{sim}}|>0,\\[6pt]
0, & \text{otherwise},
\end{cases}
\end{equation}
where $\operatorname{sim}_{j_t}$ is a tool-specific consistency metric. The total reward is
\begin{equation}
R_{\mathrm{T2S}}
=
R_{\mathrm{ans}}+\beta R_{\mathrm{obs}},
\end{equation}
where $R_{\mathrm{ans}}$ measures final-answer correctness and $\beta$ balances the two signals. Unlike answer-only RL, this objective provides direct supervision for reproducing tool observations, even when the final answer reward is sparse.

\paragraph{Privileged On-Policy Self-Distillation.}
Observation consistency supervises the predicted observations but does not fully capture how correct observations influence subsequent reasoning. We therefore introduce OPSD, using the current student model under privileged observation conditions as the teacher.

For a student-generated prefix $h_t=\hat{\tau}_{<t}$, the privileged branch receives the actual tool observations $O_{\leq k(t)}$, whereas the student conditions on its original input and generated prefix. We apply distillation to the self-simulated portions of the trajectory:
\vspace{-2mm}
\begin{equation} \small
\begin{aligned}
\mathcal{L}_{\mathrm{OPSD}}
=
&\mathbb{E}_{\hat{\tau}\sim\pi_\theta}
\Bigg[
\frac{m(\hat{\tau})}{|\mathcal{M}_{\mathrm{sim}}|}
\sum_{t\in\mathcal{M}_{\mathrm{sim}}}
D_{\mathrm{KL}}\Big(
\operatorname{sg}\big[
\pi_\theta^{\mathrm{priv}}\\
&(\cdot\mid x,h_t,O_{\leq k(t)})
\big]\big\|\,
\pi_\theta(\cdot\mid x,h_t)
\Big)
\Bigg],
\end{aligned}
\end{equation}
where $m(\hat{\tau})=\mathbb{1}[\hat{y}^{\mathrm{priv}}=y^*]$ masks samples whose privileged branch fails to produce the correct answer, and $\operatorname{sg}$ stops gradients through that branch. The loss is defined as zero when no self-simulated calls occur. This objective transfers the downstream behavior induced by actual tool evidence to the student's on-policy prefixes.

The RL stage jointly optimizes the reward and privileged distillation objectives:
\begin{equation}
\mathcal{L}_{\mathrm{RL}}
=
\mathcal{L}_{\mathrm{GRPO}}(R_{\mathrm{T2S}})
+\gamma\mathcal{L}_{\mathrm{OPSD}},
\end{equation}
where $\gamma$ balances policy optimization and self-distillation and can be scheduled alongside the tool-execution curriculum. Together, observation consistency and OPSD supervise complementary aspects of tool internalization: reproducing external evidence and reasoning effectively from that evidence. As tool execution is withdrawn, the model increasingly relies on its own predicted observations, reducing the need for external tools at inference time.

\section{Experiments}
\subsection{Experimental Setup}

\paragraph{Benchmarks.}
We evaluate T2S on 13 benchmarks covering perception (DocVQA~\cite{mathew2021docvqa}, ChartQA~\cite{masry2022chartqa}, OCRBench~\cite{liu2024ocrbench}, V*~\cite{wu2023vstar}, and HR-Bench-4K/8K~\cite{wang2024hrbench}), reasoning (MathVision~\cite{NEURIPS2024_ad0edc7d}, AI2D~\cite{DBLP:conf/eccv/KembhaviSKSHF16}, and ScienceQA~\cite{NEURIPS2022_11332b6b}), general visual understanding (GQA~\cite{hudson2019gqa}, MMBench~\cite{liu2023mmbench}, and MME~\cite{fu2023mme}), and POPE~\cite{li2023pope}. We follow the official evaluation protocols and metrics for each benchmark.

\paragraph{Baselines.}
We compare T2S with representative models spanning a wide range of parameter scales. For implicit zooming methods, we include ZwZ~\cite{wei2026zwz}, which transfers region-grounded supervision to full-image inputs, as well as Imagine-OPD~\cite{imagineopd2026} and Vision-OPD~\cite{visionopd2026}, which use on-policy self-distillation to transfer privileged visual evidence into tool-free reasoning. For explicit tool-use methods, we include Pixel-Reasoner~\cite{su2025pixelreasoner}, which learns pixel-space operations through curiosity-driven RL; Thyme~\cite{zhang2026thyme}, which executes code for image manipulation; DeepEyes~\cite{zheng2025deepeyes}, which learns active visual perception by end-to-end RL; and DeepEyesV2~\cite{hong2025deepeyesv2}, which combines SFT with RL for adaptive tool use. We additionally train a vanilla GRPO baseline using the same SFT checkpoint and training configuration as T2S.


\begin{figure*}
    \centering
    \includegraphics[width=1\linewidth]{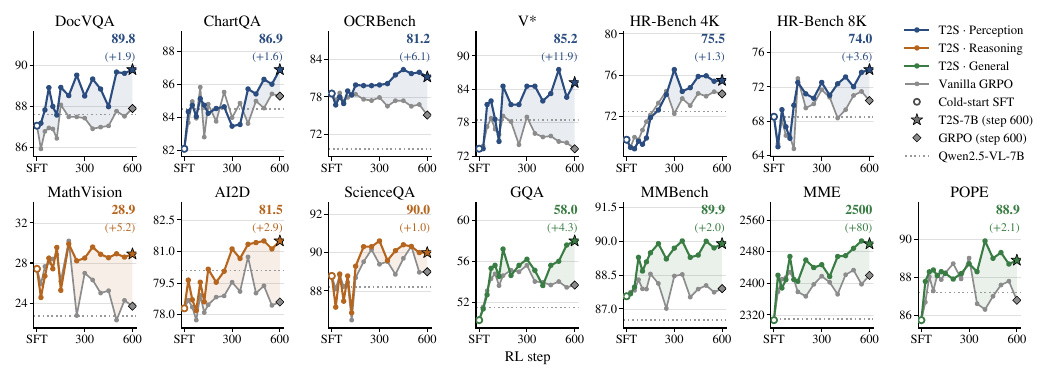}
    \vspace{-8mm}
    \caption{\textbf{Benchmark performance throughout RL steps across 13 benchmarks.} Colored and gray curves denote T2S and vanilla GRPO starting from the same cold-start checkpoint, respectively, while dotted horizontal lines indicate the base model's performance.}
    \label{fig:accuracy}
\end{figure*}

\begin{figure*}
    \centering
    \includegraphics[width=1\linewidth]{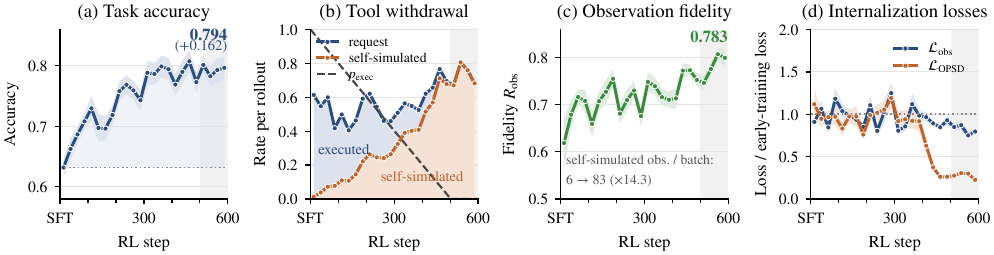}
    \vspace{-6mm}
    \caption{\textbf{Training dynamics of T2S.} (a) Task accuracy improves throughout RL. (b) External tool execution is gradually annealed, while the proportion of self-simulated observations increases. The dashed line indicates the tool-execution probability. (c) Observation fidelity improves despite the growing number of self-simulated observations. (d) Observation prediction and privileged self-distillation losses, normalized by their early-training values, decrease during training.}
    \label{fig:traininig_dynamics}
\end{figure*}

\paragraph{Training Data.}
Following the filtering and augmentation procedures described in Sec.~\ref{sec:motivation} and the repairable-trajectory construction described in Sec.~\ref{sec:method}, we construct 10K SFT samples for cold-start training and a pool of 20K candidate samples for RL. Both pools follow the same source: structural visual QA data from TextVQA~\cite{singh2019textvqa}, ChartQA~\cite{masry2022chartqa}, InfographicVQA~\cite{mathew2022infographicvqa}, and A-OKVQA~\cite{schwenk2022aokvqa}; general VQA data from GQA with test-set overlap excluded; and LLaVA-OneVision-Data~\cite{li2024llavaonevision}. For SFT, we maintain a mixture of 70\% expert-valid and 30\% expert-invalid examples. For RL, we do not pre-filter examples by tool-observation validity. Instead, we apply pass@8 filtering, removing samples for which all eight rollouts are either correct or incorrect. The retained dataset statistics are reported in the appendix.

\paragraph{Training Configuration.}
We use Qwen2.5-VL-7B as our base model. GRPO training runs for 600 iterations per run, with 16 prompts per GPU and 8 rollouts per prompt, yielding a global batch of 64 prompts (512 rollouts) per iteration. We use an actor learning rate of $2\times10^{-6}$, a clipping coefficient of 0.2, and a KL coefficient of 0.01. We set the observation prediction loss weight, observation consistency reward weight, and OPSD loss weight to $\alpha=0.1$, $\beta=0.5$, and $\gamma=0.05$, respectively. Tool-specific execution probabilities are linearly annealed over the first 500 iterations. We select three experts: GOT-OCR 2.0 \cite{wei2024general} for OCR, DePlot \cite{liu2023deplot} for chart parsing, and Grounding DINO \cite{liu2024grounding} with cropping for visual localization. Training and evaluation experiments are conducted on two separate nodes, each equipped with four NVIDIA GH200 GPUs. Details are provided in the Appendix.


\subsection{Main Results}

\noindent\textbf{Consistent Improvement Across Benchmarks.}
As shown in Table~\ref{tab:main-exp}, T2S consistently improves over Qwen2.5-VL-7B by 6.8\%, 4.9\%, and 7.2\% across perception, reasoning, and general understanding, respectively. Gains over the matched vanilla GRPO baseline remain 5.7\%, 4.8\%, and 4.4\%, isolating the contribution of explicit tool internalization beyond data quality and conventional RL. T2S further improves over vanilla GRPO by 5.2, 2.9, and 1.0 points on MathVision, AI2D, and ScienceQA, while achieving the best results on all general multimodal understanding benchmarks and POPE. These results indicate that tool internalization strengthens general visual understanding while preserving, and often improving, downstream reasoning.

\noindent\textbf{Superior Performance over Implicit Zooming Methods.}
T2S outperforms the strongest implicit zooming baseline on 9 of 13 benchmarks despite its smaller backbone. The largest gains appear on structured perception tasks, reaching 86.9 on ChartQA and 81.2 on OCRBench, respectively 12.2 and 5.1 points above Imagine-OPD-8B. This highlights the advantage of internalizing heterogeneous expert capabilities beyond regional inspection. Smaller gains on V* and HR-Bench is consistent with the information-bound regime identified in Sec.~\ref{sec:motivation}, where fine-grained evidence lost during encoding cannot be fully recovered through internalization.

\noindent\textbf{Comparison with Explicit Tool-Use Methods.}
Without external tool execution, T2S matches or surpasses most of the strongest explicit tool-use baselines. This supports our motivation that part of the benefit of external tools can be absorbed into the model, while conventional tool-use RL does not explicitly optimize this internalization. Nevertheless, explicit methods retain advantages on certain tasks (OCRBench and V*), where external observations may provide additional information unavailable to the model itself.

\begin{figure}
    \centering
    \includegraphics[width=1\linewidth]{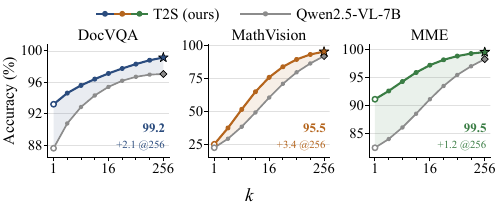}
    \vspace{-6mm}
    \caption{\textbf{Pass@k performance of T2S and its base model.} We compare T2S with Qwen2.5-VL-7B on DocVQA, MathVision, and MME under increasing sampling budgets. Annotations indicate T2S accuracy and its gain over the base model at \(k=256\).}
    \label{fig:pass_at_k}
\end{figure}

\subsection{Analysis}

\noindent\textbf{Progressive Observation Internalization.}
As shown in Fig.~\ref{fig:accuracy}, T2S improves consistently throughout RL, despite early fluctuations as the policy adapts to self-simulated observations. After roughly 200 steps, training stabilizes, with observation fidelity increasing to 0.783 despite a 14.3$\times$ increase in self-simulated observations (Fig.~\ref{fig:traininig_dynamics}(b,c)). Meanwhile, decreasing observation prediction and OPSD losses indicate progressively improved observation reconstruction and observation-conditioned reasoning (Fig.~\ref{fig:traininig_dynamics}(d)). Notably, accuracy continues to improve after step 500, when actual observations are fully withheld from the policy, demonstrating continued refinement of decisions based on self-generated evidence.

\noindent\textbf{Inference-Time Adaptation.}
Tool requests persist even after execution is fully annealed (Fig.~\ref{fig:traininig_dynamics}(b)), reflecting that T2S internalizes tool capabilities without explicitly eliminating the learned request protocol, which serves to specify perceptual subgoals. At inference, disabling external execution and removing expert-specific prompt guidance incurs less than a 1.6\% relative drop in average accuracy. The remaining gap can be avoided by retaining prompt-level expert guidance or potentially mitigated through a brief tool-free adaptation stage, which we leave for future work.

\noindent\textbf{Capability Acquisition Beyond Elicitation.}
As shown in Fig.~\ref{fig:pass_at_k}, T2S maintains an advantage over the base model as the sampling budget increases, providing evidence that tool internalization leads to genuine capability gains beyond improved access to existing capabilities. The widening gap on MathVision further suggests that internalized structured expert information may benefit downstream reasoning beyond visual perception, although disentangling these contributions remains future work.

\noindent\textbf{Beyond Privileged Distillation.} We additionally compare against a privileged-distillation baseline trained on the same 30K examples, using the same visual experts organized by Qwen3-VL-235B-A22B and an equivalent total training budget. Privileged distillation achieves an average score of 73.0, below T2S's 74.3 (Table~\ref{tab:ablation}). While it slightly outperforms T2S on reasoning, T2S achieves stronger perception and general understanding, suggesting that privileged on-policy distillation alone is insufficient to fully internalize tool capabilities without the complementary observation-consistency supervision and progressive tool scheduling.

\subsection{Ablation Study}

\begin{figure}
    \centering
    \includegraphics[width=1\linewidth]{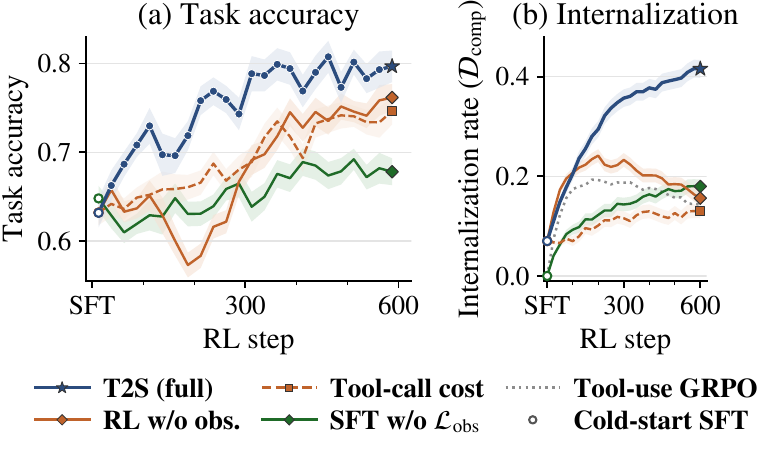}
    \vspace{-6mm}
    \caption{\textbf{Effects of training strategies on task performance and tool internalization.} (a) Task accuracy and (b) internalization rate on computation-bound examples throughout RL. }
    \label{fig:ablation_dynamics}
    
\end{figure}

Table~\ref{tab:ablation} shows that removing cold-start observation prediction reduces the average score from 74.3 to 71.5 and observation fidelity from 0.783 to 0.486. During RL, removing either observation consistency or privileged self-distillation degrades both performance and internalization, while removing both further drops the average score to 72.1 and internalization rate to 0.16, confirming their complementary roles. Progressive annealing is also essential: hard cutoff and tool-call cost reduce internalization rates to 0.27 and 0.13, respectively. Fig.~\ref{fig:ablation_dynamics} further confirms that full T2S achieves the strongest performance and most sustained internalization. Additional ablations of $\alpha$, $\beta$, and $\gamma$ are provided in the Appendix.

\input{tab/ablation}

\section{Conclusion}

We reveal that visual tool capabilities can be internalized, but conventional tool-use RL leaves this process incidental and unstable, while visual-encoding recoverability constrains its extent. To explicitly enable this process, we propose Tool-to-Self (T2S), a scaffolded RL framework that internalizes heterogeneous tool capabilities through observation supervision and progressive tool withdrawal. Across 13 benchmarks, T2S achieves leading tool-free performance, demonstrating the effectiveness of explicit tool internalization and capability acquisition beyond mere elicitation.



%% file: tab/main-exp.tex
\begin{table*}[t]
\centering
\caption{\textbf{Performance comparison across visual perception, reasoning, and general understanding benchmarks.} All results are reproduced under the same evaluation environment following the official protocols of each benchmark and baselines. MME is reported on its original score scale. “–” indicates reference values were unavailable or accurate reproduction could not be achieved.}
\label{tab:main-exp}
\setlength{\tabcolsep}{3.5pt}
\renewcommand{\arraystretch}{0.92}
\small
\resizebox{\textwidth}{!}{%
\begin{tabular}{lccccccccccccc}
\toprule
\multirow{2}{*}{\textbf{Model}}
& \multicolumn{6}{c}{\textbf{Perception}}
& \multicolumn{3}{c}{\textbf{Reasoning}}
& \multicolumn{3}{c}{\textbf{General}}
& \multirow{2}{*}{\textbf{POPE}} \\
\cmidrule(lr){2-7}
\cmidrule(lr){8-10}
\cmidrule(lr){11-13}
& DocVQA & ChartQA & OCRBench & V$^*$ & HR-4K & HR-8K
& MVision & AI2D & SciQA
& GQA & MMBench & MME
& \\
\midrule

\rowcolor{proprietarybg}

\rowcolor{proprietarybg}
\multicolumn{14}{c}{\textit{Open-Source Base Models}} \\
\midrule

Qwen2.5-VL-7B           
& 87.6 & 84.5 & 69.7 & 78.5 & 72.5 & 68.5 
& 22.7 & 80.1 & 88.2 
& 51.5 & 86.5 & 2309 
& 87.2 \\

Qwen3-VL-4B             
& 87.5 & 79.9 & 69.2 & 81.7 & 75.1 & 74.2 
& 17.9 & 79.7 & 92.6 
& 54.0 & 85.8 & 2265 
& -- \\

Qwen3-VL-8B             
& 88.8 & 82.8 & 72.7 & 82.2 & 76.2 & 72.4 
& 23.0 & 80.3 & 94.7 
& 53.5 & 87.3 & 2573 
& 89.6 \\

InternVL3-8B            
& 85.6 & 83.6 & 66.4 & 74.9 & 71.0 & 62.0 
& 24.2 & 81.7 & 98.1 
& 53.2 & 87.5 & 2495 
& 90.3 \\

\midrule
\rowcolor{reasoningbg}
\multicolumn{14}{c}{\textit{Reasoning with Tool Methods (Explicit tool-use methods)}} \\
\midrule
Pixel-Reasoner-7B 
& 86.9 & 76.2 & 77.5 & 84.3 & 73.0 & 63.6 
& 26.3 & 79.7 & \textbf{90.3} 
& 50.1 & 88.2 & 2373 
& 85.5 \\

Thyme-7B          
& 83.0 & 85.1 & \underline{86.3} & 82.2 & 77.0 & 71.8 
& \underline{27.6} & 81.0 & \textbf{90.3} 
& \underline{55.2} & \underline{89.6} & 2341 
& 85.6 \\ 

Mini-o3-7B        
& 83.8 & \underline{85.5} & 78.9 & \textbf{88.2} & 75.5 & 68.5 
& 25.7 & 75.9 & 75.7 
& \underline{55.2} & 85.5 & 2336 
& \underline{88.1} \\


DeepEyes-7B       
& 86.1 & 69.4 & 79.8 & 84.3 & 73.9 & 69.8 
& 26.6 & 80.2 & 79.3 
& 53.2 & 46.0 & 2375 
& 88.0 \\

DeepEyesV2-7B     
& -- & 72.2 & \textbf{88.2} & 81.8 & 77.9 & 73.8 
& 28.9 & -- & -- 
& 53.2 & 87.5 & 2375 
& -- \\

\midrule
\rowcolor{reasoningbg}
\multicolumn{14}{c}{\textit{Reasoning with Internal Tool Methods (Implicit zooming methods)}} \\
\midrule
ZwZ-7B 
& 86.4 & 71.4 & 64.6 & \underline{88.0} & 75.4 & 72.9 
& 25.0 & \textbf{81.7} & 87.4 
& 49.5 & 88.3 & 2199 
& 88.6 \\


Vision-OPD-4B       
& \underline{89.0} & 79.2 & 80.8 & 74.1 & 73.6 & 72.8 
& \underline{27.6} & -- & --
& 53.8 & 78.8 & 2470 
& 86.7 \\

Imagine-OPD-4B        
& 88.9 & 73.6 & 74.9 & 84.3 & 74.6 & 72.6 
& 21.4 & 77.3 & 90.1 
& 38.5 & 88.1 & 2480 
& 86.8 \\

Imagine-OPD-8B       
& 89.1 & 74.7 & 76.1 & 86.9 & 74.2 & \textbf{74.4} 
& 26.8 & 78.8 & 90.0 
& 48.2 & 87.9 & 2420 
& 87.5 \\

\midrule
\rowcolor{ourmodelsbg}
\multicolumn{14}{c}{\textit{Our Tool-to-Self Method}} \\
\midrule

\textbf{Vanilla GRPO} 
& 87.9 & 85.3 & 75.1 & 73.3 & 74.2 & 70.4 
& 23.7 & 78.6 & 89.0 
& 53.7 & 87.9 & 2420 
& 86.8 \\

\rowcolor{ourrowbg}
\textbf{T2S-7B} 
& \textbf{89.8} & \textbf{86.9} & 81.2 & 85.2 & \textbf{75.5} & \underline{74.0} 
& \textbf{28.9} & \underline{81.5} & \underline{90.0} 
& \textbf{58.0} & \textbf{89.9} & \textbf{2500} 
& \textbf{88.9} \\


$\triangle$ (\textit{vs} Vanilla GRPO) 
& +1.9 & +1.6 & +6.1 & +11.9 & +1.3 & +3.6 
& +5.2 & +2.9& +1.0 
& +4.3 & +2.0 & +80 
& +2.1 \\

\bottomrule
\end{tabular}%
}
\end{table*}

%% file: tab/ablation.tex

\begin{table}[t]
  \centering
  \caption{\textbf{Ablations of T2S.} Perc., Reas., and Gen.\ are category averages over
  the benchmarks of Table~\ref{tab:main-exp} (excludes MME and POPE). Int.\ rate is the
  internalization rate on $\mathcal{D}_{\mathrm{comp}}$ at step 600 relative to the
  vanilla-GRPO checkpoint (Sec.~\ref{sec:motivation}); $R_{\mathrm{obs}}$ is the fidelity
  of self-simulated observations over steps 500--600.}
  \label{tab:ablation}
  \renewcommand{\arraystretch}{0.92}
  \setlength{\tabcolsep}{3.2pt}
  \resizebox{\linewidth}{!}{%
  \begin{tabular}{lcccccc}
    \toprule
    Variant & Perc. & Reas. & Gen. & Avg. & Int.\ rate & $R_{\mathrm{obs}}$ \\
    \midrule
    Vanilla GRPO & 77.7 & 63.8 & 70.8 & 70.8 & -- & -- \\
    \midrule
    \rowcolor{ourrowbg}
    T2S (full) & \textbf{82.1} & \textbf{66.8} & \textbf{74.0} & \textbf{74.3} & \textbf{0.42} & \textbf{0.783} \\
    \midrule
    \multicolumn{7}{l}{\emph{\textbf{Cold-start SFT}}} \\
    \quad w/o $\mathcal{L}_{\mathrm{obs}}$ (tool calls only) & 78.6 & 64.4 & 71.4 & 71.5 & 0.18 & 0.486 \\
    \midrule
    \multicolumn{7}{l}{\emph{\textbf{Internalization signals in RL}}} \\
    \quad w/o $R_{\mathrm{obs}}$ & 80.0 & 65.9 & 73.0 & 73.0 & 0.30 & 0.618 \\
    \quad w/o $\mathcal{L}_{\mathrm{OPSD}}$ & 80.9 & 66.3 & 73.5 & 73.6 & 0.35 & 0.721 \\
    \quad w/o obs. & 79.2 & 65.0 & 72.0 & 72.1 & 0.16 & 0.552 \\
    \midrule
    \multicolumn{7}{l}{\emph{\textbf{Tool annealing schedule}}} \\
    \quad Cosine annealing & 81.6 & 66.5 & 73.6 & 73.9 & 0.39 & 0.768 \\
    \quad Hard cutoff (step 250) & 79.6 & 65.4 & 72.5 & 72.5 & 0.27 & 0.665 \\
    \quad Tool-call cost & 78.9 & 64.9 & 71.8 & 71.9 & 0.13 & -- \\
    \midrule
    \rowcolor{proprietarybg}
    Privileged Distillation & 80.9 & \textbf{67.2} & 71.0 & 73.0 & -- & -- \\
    \bottomrule
  \end{tabular}}
\end{table}
